\documentclass[10pt,twocolumn,letterpaper]{article}

\usepackage[pagenumbers]{cvpr} 

\usepackage{booktabs}
\usepackage{xcolor}
\usepackage{colortbl}
\usepackage{multirow}

\definecolor{oursblue}{HTML}{D9E7F4}
\definecolor{bestred}{HTML}{D7263D}
\definecolor{gaingreen}{HTML}{2E8B57}
\definecolor{venuegray}{HTML}{8C8C8C}
\newcommand{\best}[1]{\textcolor{bestred}{#1}}

\newcommand{\gain}[1]{\textcolor{gaingreen}{#1}}

\newcommand{\reftable}{\setlength{\aboverulesep}{0pt}\setlength{\belowrulesep}{0pt}\renewcommand{\arraystretch}{1.18}\setlength{\tabcolsep}{4pt}}

\definecolor{cvprblue}{rgb}{0.21,0.49,0.74}
\usepackage[pagebackref,breaklinks,colorlinks,allcolors=cvprblue]{hyperref}

\def\confName{CVPR}
\def\confYear{2027}

\title{Match Before You Merge: Region Evidence as a Spatial Inductive Bias\\for Cross-View Localization}

\author{%
Jialang Liu\textsuperscript{1}\quad
Tianfang Tong\textsuperscript{2}\quad
Le Zhou\textsuperscript{3}\quad
Yingmei Wei\textsuperscript{1}\quad
Yanming Guo\textsuperscript{1,*}\\[4pt]
{\small\textsuperscript{1}National Key Laboratory for Big Data and Decision,}\\
{\small National University of Defense Technology (NUDT), Hunan, China}\\
{\small\textsuperscript{2}Changsha University of Science and Technology (CSUST), Hunan, China}\\
{\small\textsuperscript{3}Shanghai Jiao Tong University, Shanghai, China}\\[4pt]
{\small Jialang Liu: \texttt{liu\_1999@nudt.edu.cn}\quad
Tianfang Tong: \texttt{202403060216@csust.edu.cn}}\\
{\small Le Zhou: \texttt{zhoule116@sjtu.edu.cn}\quad
Yingmei Wei: \texttt{weiyingmei@nudt.edu.cn}}\\
{\small Yanming Guo\textsuperscript{*}: \texttt{guoyanming@nudt.edu.cn}}\\[2pt]
{\small\textsuperscript{*}Corresponding author.}
}

\begin{document}
\twocolumn[{%
\renewcommand\twocolumn[1][]{#1}%
\maketitle
\begin{center}
\centering
\captionsetup{type=figure}
\includegraphics[width=\textwidth]{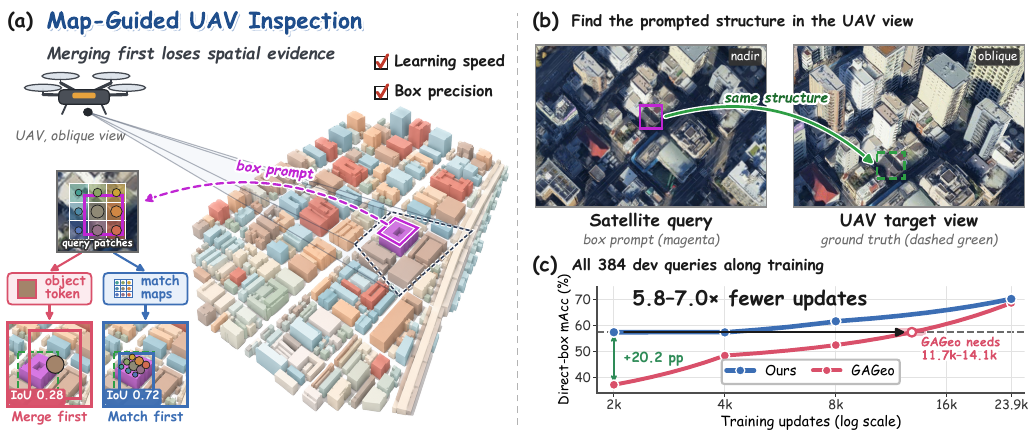}
\vspace{-1.2em}
\captionof{figure}{\textbf{Match before you merge.}
(a)~Map-guided UAV inspection. A box on the satellite image (magenta)
marks the building to localize in the oblique UAV view. The inset shows
the prompted patches, with colour indicating position and dot size
indicating the coverage weight $p_i$. GAGeo regresses a box from its
updated object token, while our head matches each patch to the UAV grid
before combining the match maps. The schematic overlays their measured
boxes at 2k updates (IoU 0.28 and 0.72, respectively). The equal-capacity
pooled control reaches IoU 0.73 on this query (supplementary Fig.~8).
Across all 384 development queries, matching first improves direct-box
mAcc by $8.15$\,pp over that control (\cref{tab:ablation}).
(b)~The satellite and UAV images for the same query (Tokyo 20282), with
the ground-truth box dashed in green.
(c)~Direct-box mAcc on dev384 for the paired strict runs (seed 2026).
Our head reaches 57.40\% at 2k updates (dashed line). Interpolation on
the GAGeo learning curves places the same level at 11.7k--14.1k updates
(shaded band), a $5.8$--$7.0\times$ update ratio (\cref{tab:budget}).}
\label{fig:concept}
\end{center}
\vspace{0.6em}

}]
\begin{abstract}
Satellite-to-UAV object localization uses a box on a satellite image to specify the building to locate in an oblique UAV view. Changes in visible surfaces, scale, and orientation make it difficult to place the box precisely even when the building is correctly identified. We introduce a \emph{match-before-merge} head that preserves patch-level spatial evidence for center prediction. Each prompted satellite patch first produces a normalized matching distribution over the UAV grid. The head combines these distributions in proportion to each patch's coverage by the query box. Each target-grid location casts a center vote with a learned bounded offset, and the head predicts the box center as the expectation of these votes under the combined distribution. After $2{,}000$ updates, our head reaches $57.40\%$ mean direct-box accuracy (mAcc) on 384 CMA-Loc development queries. This is $8.15$ percentage points above a pooled control with the same head capacity and initialization, isolating the effect of aggregation order. Relative to a paired direct-box baseline, our head reaches the same accuracy in approximately one-sixth as many updates, as estimated by interpolation on the strict learning curve, at comparable cost per update. At the final checkpoint, our head improves direct-box mAcc by $1.34$ percentage points on the six-city validation set and $1.30$ points in a held-out city. On the six-city set, the gain is $0.24$ points at an intersection-over-union threshold of $0.50$ and $3.72$ points at $0.90$, showing that the final gains concentrate where precise box placement matters most.
\end{abstract}
\section{Introduction}
\label{sec:intro}

A box on a satellite image marks a structure for a UAV to inspect. Where should that box land in the UAV's oblique view? The two views can show different surfaces of the same building, at different scales and orientations. Precisely placing the box is a challenge in its own right. At an intermediate GAGeo checkpoint, replacing the direct box's center with the center of its own mask-derived box improves direct-box mAcc by $13.52$ percentage points. Replacing only its width and height improves it by $2.34$ points (see the supplementary material). This gap motivates a closer look at how the model predicts the center.

Cross-view localization uses spatial structure to bridge the viewpoint gap, through polar transforms~\cite{shi2019spatial}, orientation maps~\cite{liu2019lending}, and geometric masks~\cite{lentsch2023slicematch}. GAGeo~\cite{gageo} models local and global cross-view interactions and predicts the box from an updated object token. We focus on the final step that turns those features into a location. \textbf{Can the center readout learn faster if it matches the prompted patches before merging their evidence?}

\begin{figure*}[t]
\centering
\includegraphics[width=\textwidth]{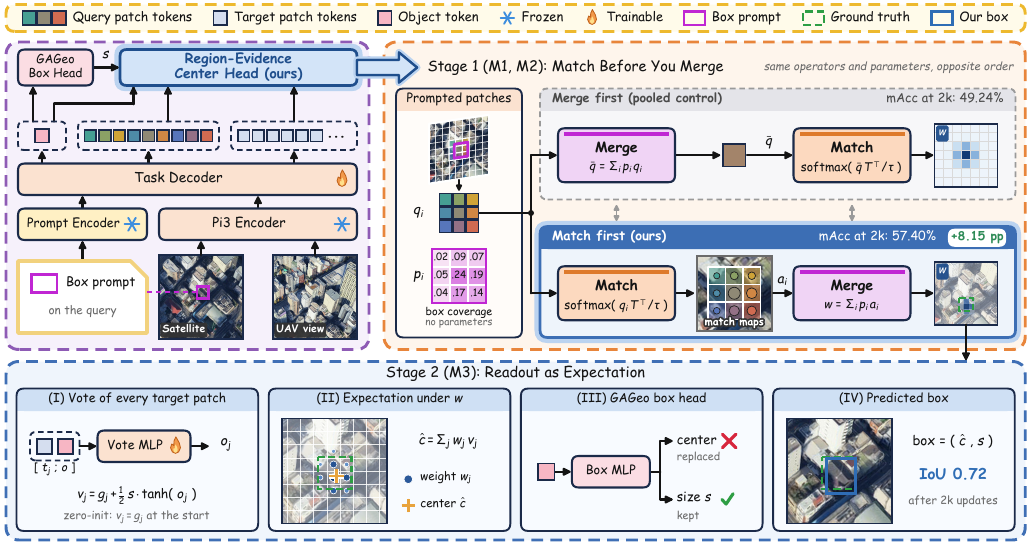}
\caption{\textbf{Method overview} on the query of \cref{fig:concept}. \textbf{Left}: the GAGeo backbone is shared and unchanged. GAGeo's box head reads only the object token; our head also reads the query and target patch tokens and keeps the box size $s$. \textbf{Stage 1}: the same two operators in two orders. The pooled control merges the nine prompted query patches into one token with the coverage weights $p_i$ and matches once. Our head first matches every patch $q_i$ to the $37{\times}37$ target grid (M1), then merges the nine match maps $a_i$ with the same weights into the region evidence $w$ (M2). With equal parameters and initialization, matching first adds 8.15\,pp of direct-box mAcc on dev384 at 2{,}000 updates (\cref{tab:ablation}). \textbf{Stage 2} (M3): every target patch casts a vote $v_j$, and the center $\hat{c}$ is the expectation of the votes under $w$. The weights $p_i$, the evidence $w$, and the boxes are measured after 2{,}000 updates (IoU 0.72); the match maps and the backbone blocks are schematic.}
\label{fig:framework}
\end{figure*}

We introduce a \emph{match-before-merge} center head for satellite-to-UAV localization (\cref{fig:framework}). Each prompted satellite patch first produces a normalized matching distribution over the UAV grid. We mix these distributions in proportion to each patch's coverage by the prompt box and predict the center as a spatial expectation over target-grid votes. Each vote starts at its grid center and can learn a bounded offset. The resulting head retains the contribution of each prompted patch until its match is expressed in the coordinates of the target view.

\noindent\textbf{Contributions.}
\begin{enumerate}
    \item \textbf{Aggregation order as a spatial inductive bias.}
    We compare the proposed head with a pooled variant that uses the same capacity and initialization but merges patch features before matching. At $2{,}000$ updates on dev384, matching first raises direct-box mAcc from $49.24\%$ to $57.40\%$, a gain of $8.15$\,pp (\cref{tab:ablation}).
    \vspace{2pt}

    \item \textbf{Fewer updates to reach the same localization quality.}
    With the same backbone, data, and training schedule, the proposed head reaches $57.40\%$ direct-box mAcc at $2{,}000$ updates, versus $37.16\%$ for GAGeo. Interpolation on GAGeo's learning curves puts its crossing of $57.40\%$ at $11.7$k--$14.1$k updates, a $5.8$--$7.0\times$ update ratio at comparable per-update cost ($4.58$ vs.\ $4.53$\,s; \cref{fig:concept}).
    \vspace{2pt}

    \item \textbf{More precise boxes at the final checkpoint.}
    After $23{,}937$ updates, the direct-box gain is $+1.34$\,pp on the full six-city validation set and $+1.30$\,pp on held-out Los Angeles. The gains concentrate at strict IoU thresholds, reaching $+3.72$\,pp at IoU~$0.90$ on the full validation set. Our direct boxes also outperform GAGeo's mask-derived boxes on both sets.
\end{enumerate}
\section{Related work}
\label{sec:related}

\noindent\textbf{Cross-view geo-localization.} Matching ground-level queries against aerial reference imagery has been studied as retrieval~\cite{workman2015cvusa,vo2016localizing}, where global descriptors are compared and spatial structure enters through explicit geometric priors: per-pixel orientation maps~\cite{liu2019lending} and orientation supervision~\cite{vo2016localizing}, polar transforms aligning the two viewpoints~\cite{shi2019spatial}, and aerial-to-ground layout translation~\cite{zhai2017predicting}. Benchmarks moved the field from aligned pairs to realistic unaligned settings~\cite{zhu2021vigor}, and recent work learns the spatial configuration itself, via transformer backbones~\cite{zhu2022transgeo}, disentangled geometric layout descriptors~\cite{zhang2023geodtr}, or hard-negative contrastive training~\cite{deuser2023sample4geo}. Drone-based geo-localization extends retrieval to UAV views and poses satellite-to-UAV navigation as a task of its own~\cite{zheng2020university}. SliceMatch~\cite{lentsch2023slicematch} constructs a pose heatmap by comparing a ground descriptor with pose-dependent aerial descriptors. We study how preserving per-patch matching distributions until aggregation affects the optimization budget for prompted satellite-to-UAV localization on CMA-Loc~\cite{gageo}.
\vspace{2.5pt}

\noindent\textbf{Local interaction before aggregation.} ColBERT~\cite{khattab2020colbert} scores query--document relevance
by summing token-level maximum similarities, while FILIP~\cite{yao2021filip} uses token-wise maximum similarity for
image--text contrastive pretraining. We construct a spatial
distribution for localization by normalizing each query patch's
matches over the target grid and mixing them with query-box
coverage weights. An equal-capacity pooled control measures
the effect of aggregation order.
\vspace{2.5pt}

\noindent\textbf{Coordinate readout as expectation.} Using a spatial distribution to obtain coordinates is established: soft-argmin stereo regression~\cite{kendall2017gcnet} and integral pose regression~\cite{sun2018integral} connect spatial responses with differentiable coordinate prediction. Our contribution precedes the readout; it concerns how region-conditioned spatial evidence is \emph{constructed}, and the expectation readout then exploits that structure with a zero-initialized, calibrated start.
\vspace{2.5pt}

\noindent\textbf{Cross-view object localization.} Object-level cross-view localization was posed on the CVOGL benchmark~\cite{sun2023detgeo}, where an object marked in a ground or drone image is localized in a satellite image. Later methods add location enhancement with multi-head cross attention~\cite{huang2025ocgnet} or predict object masks instead of boxes~\cite{zhang2025trogeo}. These benchmarks localize the object in the satellite view; we study the reverse direction, from satellite to UAV. GAGeo~\cite{gageo} combines prompt and task tokens with local/global cross-view interactions and regresses the target box from an updated object token. Our modification concerns the final center readout, explicitly combining query-region patch-to-target distributions. The pooled control swaps only the readout inside our head; the GAGeo baseline keeps its original decoder and is retrained in the satellite-to-UAV direction. OSGeo~\cite{osgeo} addresses oriented objects via rotated boxes and an orientation-sensitive head; HiSymGeo~\cite{hisymgeo} spatially averages the enhanced query map before multi-scale fusion compares the global descriptor with reference features. GAGeo, the dataset's public state of the art, is our primary baseline, and internal control variants isolate the design factors.
\vspace{2.5pt}

\noindent\textbf{Inductive bias and learning speed.} 
Spatial inductive biases encode architectural structure to improve
learning~\cite{battaglia2018relational,cohen2016group}.
We examine how aggregation order affects optimization efficiency
in cross-view localization.

\section{Method}
\label{sec:method}

\subsection{Problem setup}
Input: a satellite query image, a query box on it that marks the target structure (normalized xyxy), and an oblique UAV target image. Output: the target box in the UAV view (normalized center + size). Both images enter at 518\,px, so the target grid has $37{\times}37$ patches. Backbone and features follow GAGeo: query/target patch features $Q \in \mathbb{R}^{B \times N_q \times C}$, $T \in \mathbb{R}^{B \times N_t \times C}$ ($C{=}2048$), object token $O \in \mathbb{R}^{B \times 1 \times C}$, and box size $s \in \mathbb{R}^{B \times 1 \times 2}$, the width and height predicted by GAGeo's box head. \textbf{We replace only the final center readout}, a single change point; backbone, data, budget, initialization, and schedule are shared with the GAGeo baseline.

\subsection{Region evidence center head}
\Cref{fig:framework} shows the head and its place after the shared backbone.

\textbf{M1: Local matching distributions (match first).}
Three linear projections to $d{=}128$ (target/query/object), L2-normalized with explicit FP32 contraction. Each query patch $i$ forms a normalized distribution over target patches,
\begin{equation}
a_i = \mathrm{softmax}\!\left(q_i\, T^{\top} / \tau\right),
\end{equation}
with learnable $\tau$ sigmoid-bounded to $(0.01, 1.00)$, initialized at $0.07$. Spatial correspondence is preserved per patch as a distribution, never compressed into a single vector.

\textbf{M2: Coverage-weighted evidence mixing (merge by belonging).}
Fractional box coverage $c_i \in [0,1]$ of the query box over each query patch (analytic, zero parameters) gives mixing weights $p_i = c_i / \sum_k c_k$. Region spatial evidence is the normalized target-grid distribution
\begin{equation}
w = \sum_i p_i \cdot a_i .
\end{equation}

\textbf{M3: Center voting readout (readout as expectation).}
\begin{equation}
\hat{c} = \sum_j w_j \cdot \left(g_j + \tfrac{1}{2}\, s \cdot \tanh(o_j)\right),
\end{equation}
where $g_j$ are target-grid centers and $o_j = \mathrm{MLP}([t_j; o])$ a bounded offset. The zero-initialized final layer makes the initial prediction already the coverage-weighted spatial expectation: the readout is calibrated at optimization step zero. The size $s$ is the width and height predicted by GAGeo's box head, so the readout's learned parameters act on the box center, the component that dominates GAGeo's localization error (supplementary material).

\subsection{Control variants}
Ablations switch a single \texttt{mode} of the same head: \emph{reciprocal} (adds a backward-pointing quality term), \emph{spatial} (object-token-only scoring, no per-patch match), and \emph{pooled} (aggregate-then-match, the core ablation on aggregation timing). Forward, reciprocal, and pooled share 819{,}971 head parameters and identical initialization tensors; the spatial head has 557{,}699. A training-free control, the \emph{mask-center swap}, moves the center of GAGeo's direct box to the center of its predicted mask and keeps the direct box size.

\subsection{Why the readout lowers the optimization cost}
Three properties of our readout can reduce the budget needed to reach a given localization quality; only the first depends on matching first, and the pooled control shares the other two.
\textbf{Dense spatial credit assignment.} Pooled readout yields one gradient direction per sample; match-first exposes every patch's matching distribution to the localization loss through the mixing weights, a short gradient path from error to local features.
\textbf{Calibrated start.} Zero-initialized offsets make the initial readout a sensible spatial expectation, while the baseline decoder must first learn the readout structure and then the correspondence. The lead is therefore largest at the earliest budgets.
\textbf{Metric-aligned parameterization.} The readout is isomorphic to the box-center metric with few free parameters, so the baseline approaches it given enough budget. The gap narrows with budget and stays positive ($+20.2$, $+8.9$, $+9.2$, and $+1.5$\,pp at 2k, 4k, 8k, and 23{,}937 updates on dev384).
\cref{sec:curve} returns to these properties with the measured learning curve.

\begin{figure*}[t]
\centering
\includegraphics[width=\textwidth]{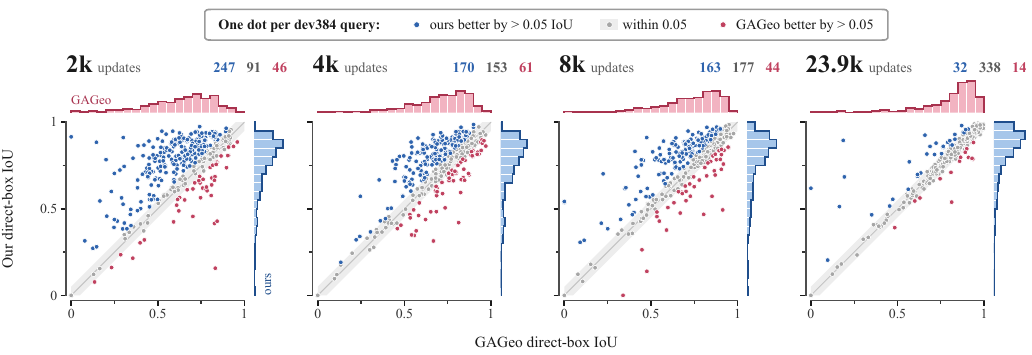}
\\caption{\textbf{Per-query localization along training}
(dev384, strict profile, seed 2026).
Each dot compares GAGeo's direct-box IoU ($x$) with ours ($y$)
at the same budget. Marginal histograms use 0.05-wide bins and
a shared count scale. Blue/rose dots indicate an advantage
exceeding 0.05 IoU for ours/GAGeo; gray dots differ by at most 0.05.
Panel counts follow these colors.
At 2k updates, ours leads on 247 queries and trails on 46
(median IoU 0.80 vs 0.67); at 23{,}937 updates, 338 queries lie
in the gray band. Learning curves and matched-accuracy budgets
appear in \cref{fig:concept}(c) and \cref{tab:budget}.}
\label{fig:curve}
\end{figure*}

\begin{figure*}[t]
\centering
\includegraphics[width=\textwidth]{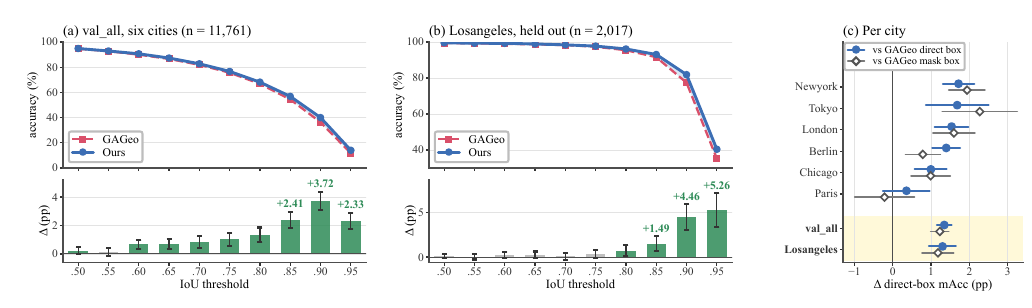}
\caption{\textbf{Where the converged lead comes from} (23{,}937 updates). (a, b) Direct-box accuracy at each IoU threshold of the mAcc grid on val\_all and held-out Losangeles (top), and the paired difference, ours minus GAGeo, with 95\% source-grid cluster-bootstrap intervals (bottom; green: interval excludes zero). The gains concentrate at strict IoU thresholds. (c) Per-city difference in direct-box mAcc against GAGeo's direct box (filled) and against its mask-derived box (open), with 95\% intervals; the shaded rows pool all six validation cities and give the held-out city.}
\label{fig:forest}
\end{figure*}

\section{Experiments}
\label{sec:exp}

\subsection{Evaluation protocol}
\label{sec:protocols}
\textbf{Data and baseline.}
CMA-Loc~\cite{gageo} pairs oblique UAV views rendered from Google Photorealistic 3D Tiles with nadir orthographic satellite views. We use all 95{,}742 training pairs and validate on six cities (Berlin, Chicago, London, Newyork, Paris, Tokyo). Both arms start from GAGeo's released weights and train satellite-to-UAV using its optimizer settings and losses, with heatmap and rotation losses disabled. Training lasts three epochs (23{,}937 updates, effective batch 12), with a cosine schedule and 500 warmup updates.
\textbf{Strict determinism.}
Paired runs use \emph{strict-deterministic-v1} (torch 2.5.1+cu124, deterministic algorithms, cuDNN benchmark off, FP32 contraction). Equal-capacity heads share 819{,}971 parameters and identical initialization.
\textbf{Evaluation.}
Learning curves and ablations use dev384 (64 queries per city). Final evaluation uses val\_all (11{,}761 queries, including dev384) and held-out Losangeles (2{,}017 queries), evaluated once after freezing the method and checkpoints.
\textbf{Metrics.}
Direct-box mAcc averages accuracy over IoU thresholds $0.50{:}0.05{:}0.95$. Mask-derived box mAcc evaluates boxes fitted to predicted masks; mask mIoU measures segmentation.
\textbf{Statistics.}
We report equally weighted per-sample paired differences with source-grid cluster-bootstrap intervals (20{,}000 resamples, seed 2026). Intervals capture sample variation; a pre-registered second GAGeo seed (2027; \cref{sec:curve}) assesses initialization sensitivity.

\begin{figure*}[t]
\centering
\includegraphics[width=\textwidth]{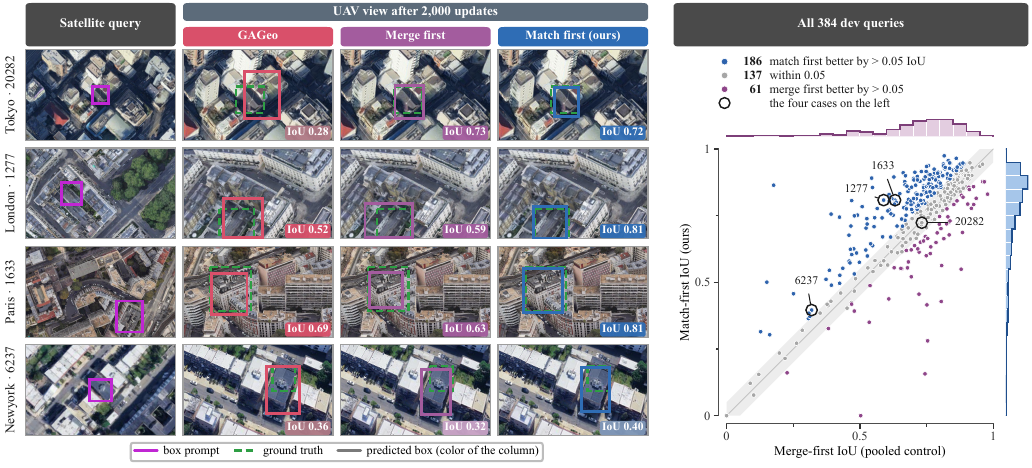}
\caption{\textbf{Aggregation order at equal capacity}
(dev384, 2{,}000 updates).
\textbf{Left}: four of 12 cases fixed before evaluation,
showing satellite prompts (magenta), UAV ground truth
(green, dashed), predictions, and IoUs.
Both orders find Tokyo 20282; matching first tightens the
London 1277 and Paris 1633 boxes; both miss Newyork 6237
at IoU 0.5.
\textbf{Right}: paired IoUs on all 384 queries.
Matching first gains over 0.05 IoU on 186 queries and loses
over 0.05 on 61 (median 0.80 vs 0.75).
Both heads share 819{,}971 parameters and initialization.
The remaining eight cases appear in the supplementary material.}
\label{fig:qual}
\end{figure*}

\begin{figure*}[!h]
\centering
\includegraphics[width=0.85\textwidth]{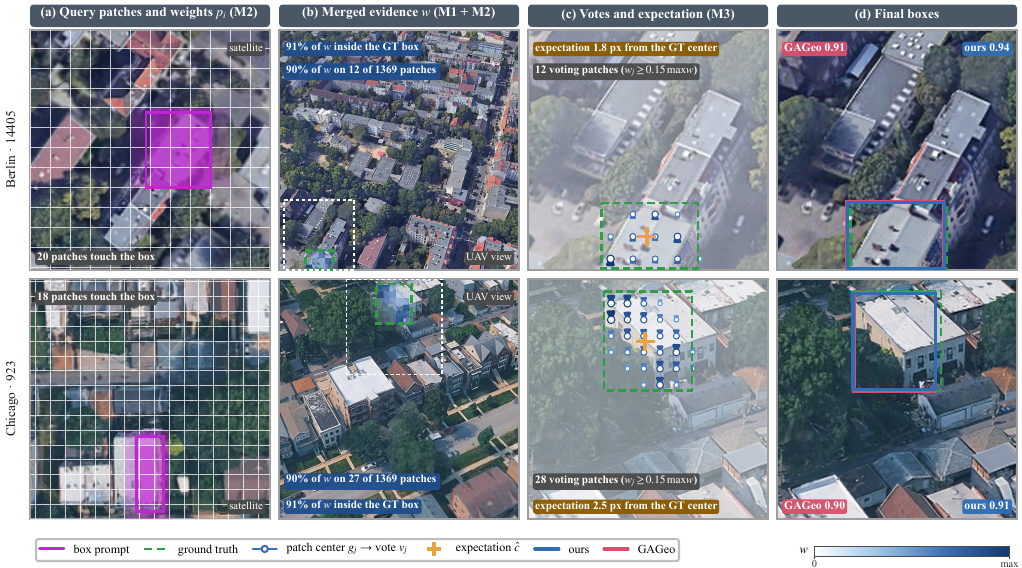}
\caption{\textbf{Inside the region-evidence head} at the final checkpoint (23{,}937 updates, seed 2026), for two of the 12 prespecified cases. (a)~Satellite query patches with their box-coverage weights $p_i$ (shade). (b)~Merged evidence $w$ over the UAV view (blue); the white window is enlarged in (c) and (d). (c)~Every patch with $w_j \geq 0.15\max w$ votes for a center $v_j$ within half a box of its own center $g_j$ (arrows, shade by weight); the amber cross is the expectation $\hat{c}$. (d)~Final boxes of ours and GAGeo with their IoUs.}
\label{fig:mechanism}
\end{figure*}

\begin{table}[t]
\centering
\caption{\textbf{Paired comparison with GAGeo~\cite{gageo} at the final checkpoint} (23{,}937 updates, strict profile, seed 2026), in percent. val\_all is the full validation set; Losangeles is a held-out city evaluated once after freezing method and checkpoints. Mask box: mask-derived box mAcc; $^\dagger$our direct box against GAGeo's mask-derived box. Intervals are 95\% paired source-grid cluster-bootstrap intervals; \best{red}: higher value; \gain{green}: interval excludes zero.}
\label{tab:main}
\reftable
\resizebox{\columnwidth}{!}{%
\footnotesize
\begin{tabular}{l|cc|cc}
\toprule
Metric & GAGeo & \cellcolor{oursblue}Ours & $\Delta$ (pp) & 95\% CI \\
\midrule
\multicolumn{5}{l}{\emph{val\_all, six cities ($n{=}11{,}761$)}} \\
Direct box & 69.05 & \cellcolor{oursblue}\best{70.40} & \gain{+1.34} & [1.15, 1.53] \\
Mask box & 69.17 & \cellcolor{oursblue}\best{69.34} & +0.17 & [$-$0.02, 0.37] \\
Mask mIoU & 80.01 & \cellcolor{oursblue}\best{80.12} & \gain{+0.11} & [0.01, 0.20] \\
Direct vs.\ mask box$^\dagger$ & 69.17 & \cellcolor{oursblue}\best{70.40} & \gain{+1.23} & [0.99, 1.46] \\
\midrule
\multicolumn{5}{l}{\emph{Losangeles, held out ($n{=}2{,}017$)}} \\
Direct box & 89.16 & \cellcolor{oursblue}\best{90.46} & \gain{+1.30} & [0.94, 1.65] \\
Mask box & 89.28 & \cellcolor{oursblue}\best{89.46} & +0.18 & [$-$0.15, 0.51] \\
Mask mIoU & 92.20 & \cellcolor{oursblue}\best{92.33} & \gain{+0.13} & [0.01, 0.26] \\
Direct vs.\ mask box$^\dagger$ & 89.28 & \cellcolor{oursblue}\best{90.46} & \gain{+1.18} & [0.76, 1.59] \\
\bottomrule
\end{tabular}}
\end{table}

\subsection{Optimization budget: the learning curve}
\label{sec:curve}
\Cref{fig:concept,fig:curve,tab:budget} connect early accuracy to optimization budget. On the strict paired curves, our direct-box mAcc is 57.40 / 57.34 / 61.61 / 70.21\% at 2k / 4k / 8k / 23{,}937 updates; GAGeo: 37.16 / 48.41 / 52.45 / 68.67. GAGeo crosses 57.40\% at ${\approx}$12{,}900 updates by piecewise-linear interpolation of the strict curve (52.45 at 8k $\rightarrow$ 68.67 at 23{,}937) and at ${\approx}$14{,}054 updates on its denser standard-profile curve (12k${=}$50.18 $\rightarrow$ 16k${=}$64.24), a 6.4--7.0$\times$ budget ratio. Ours needs 2.54 GPU-hours against 17.68; per-update cost is 4.58\,s vs 4.53\,s ($+1.1$\%) and peak memory 33.2 vs 33.1\,GB. At the final checkpoint ours keeps a $+1.54$\,pp direct-box lead (70.21 vs 68.67) and a $+0.72$\,pp mask mIoU lead (80.59 vs 79.87) on dev384.

\noindent\textbf{Second seed.}
The pre-registered GAGeo replication (seed 2027, same configuration) reaches 57.40\% at ${\approx}$11{,}700 updates (5.8$\times$) and finishes at 68.91\%, $1.30$\,pp below ours and $0.24$\,pp above seed 2026. Combined with the seed-2026 standard-profile reference, this yields the 5.8--7.0$\times$ range in \cref{fig:concept} (supplementary material).

\noindent\textbf{What carries the advantage.}
The lead is largest early ($+20.2$\,pp at 2{,}000 updates) and remains
positive at every evaluated checkpoint. At 2{,}000 updates, our
direct-box IoU exceeds GAGeo's by more than 0.05 on 247 of 384 queries
and trails it by that margin on 46; at 23{,}937 updates, 338 queries
differ by at most 0.05 (Fig.~3). This pattern is consistent with the
proposed spatial readout facilitating early learning
(\cref{sec:method}). On quarter data, both arms complete three passes
and their cosine schedules in 5{,}994 updates, reaching 65.55\% and
64.40\% mAcc (supplementary material). The smaller gap under this
setting suggests that the early advantage depends on training
progress, with repeated exposure and schedule completion both
potentially contributing.

\begin{table}[t]
\centering
\caption{\textbf{Optimization budget to matched localization quality}
(dev384). Updates and GPU-hours to reach 57.40\% direct-box mAcc,
the level our head reaches at 2{,}000 updates. GAGeo's crossings
use seed 2026's standard-profile reference and seed 2027's strict
curve (\cref{sec:curve}). Linear interpolation gives update ratios
of 7.0$\times$ and 5.8$\times$, respectively. Per-update costs use
the strict runs' resumed segments; GPU-hours are estimated from
these costs and the corresponding update counts.}
\label{tab:budget}
\reftable
\resizebox{\columnwidth}{!}{%
\begin{tabular}{l|cc|cc}
\toprule
\multirow{2}{*}{Method} & \multicolumn{2}{c|}{To 57.40\% mAcc} & \multicolumn{2}{c}{Per update} \\
 & Updates & GPU-hours & Seconds & Memory (GB) \\
\midrule
GAGeo, seed 2026 & 14{,}054 & 17.68 & 4.53 & 33.1 \\
GAGeo, seed 2027 & 11{,}682 & -- & -- & -- \\
\rowcolor{oursblue} Ours & \best{2{,}000} & \best{2.54} & 4.58 & 33.2 \\
\midrule
Budget ratio & \multicolumn{2}{c|}{\gain{7.0$\times$ / 5.8$\times$}} & \multicolumn{2}{c}{} \\
\bottomrule
\end{tabular}}
\end{table}

\subsection{Convergence}
\label{sec:asymptotic}
On the full validation set ($n{=}11{,}761$), ours reaches 70.40\% versus 69.05\% direct-box mAcc: $\mathbf{+1.34}$\,pp [$+1.15, +1.53$], positive in all six cities (range $+0.36$ Paris to $+1.72$ Newyork). Gains concentrate at strict IoU thresholds (\cref{fig:forest}a): $+0.24$\,pp at IoU 0.50, $+2.41$\,pp at 0.85, and $+3.72$\,pp at 0.90, where GAGeo localizes 36.28\% of the queries and ours 40.01\%. Our readout therefore sharpens where the box is placed, the component it acts on. Mask mIoU also rises, $+0.11$ [$+0.01, +0.20$], and the mask-derived box changes by $+0.17$ [$-0.02, +0.37$] (\cref{tab:main}).

\subsection{Out-of-distribution generalization}
\label{sec:ood}
On held-out Losangeles ($n{=}2{,}017$; 501 source-grid clusters), evaluated once after freezing method and checkpoints: direct-box mAcc 90.46\% vs 89.16\%, $\mathbf{+1.30}$\,pp [$+0.94, +1.65$]; mask-derived box $+0.18$ [$-0.15, +0.51$]; mask mIoU $+0.13$ [$+0.01, +0.26$] (\cref{fig:forest,tab:main}). The direct-readout advantage survives distribution shift and again concentrates at strict thresholds: $+4.46$\,pp at IoU 0.90 and $+5.26$\,pp at 0.95 (\cref{fig:forest}b). Mask mIoU reaches 92.33 vs 92.20, and our direct readout also exceeds GAGeo's mask-derived box, by $+1.18$\,pp [$+0.76, +1.59$] on Losangeles and in five of the six validation cities (\cref{fig:forest}c).

\subsection{Ablations at matched budget (2{,}000 updates)}
\label{sec:ablation}
\Cref{tab:ablation} isolates the design factors at 2{,}000 updates (dev384, seed 2026). Forward, reciprocal, and pooled heads share 819{,}971 parameters and identical initialization tensors; the spatial head has 557{,}699. Aggregation timing is the largest structural factor: pooling before matching costs 8.15\,pp. Per-patch matching matters (the spatial head loses 4.61\,pp), and adding a reciprocal term lowers mAcc by 2.47\,pp, so the forward-only head is both simpler and more accurate. Our head also beats a training-free fix of GAGeo that moves its direct box to the center of its predicted mask ($+3.18$\,pp). GAGeo's mask-derived box reaches 60.13\% at this budget; it comes from GAGeo's mask pathway, which our head leaves unchanged, and by convergence our direct readout exceeds it on val\_all (70.40 vs 69.17\%, $+1.23$\,pp [$+0.99, +1.46$]).

\begin{table}[t]
\centering
\caption{\textbf{Ablations at 2{,}000 updates} (dev384, seed 2026). $\Delta$ is ours minus the variant, with paired 95\% cluster-bootstrap intervals (city-equal-weighted); \gain{green}: interval excludes zero. GAGeo rows use the baseline checkpoint at the same budget; the mask-center swap moves its direct box to the center of its predicted mask without training.}
\label{tab:ablation}
\reftable
\resizebox{\columnwidth}{!}{%
\begin{tabular}{l|c|c|cc}
\toprule
Variant & Head params & mAcc (\%) & $\Delta$ (pp) & 95\% CI \\
\midrule
\rowcolor{oursblue} Ours & 819{,}971 & \best{57.40} & -- & -- \\
+ Reciprocal term & 819{,}971 & 54.92 & \gain{+2.47} & [0.48, 4.48] \\
Spatial (no per-patch match) & 557{,}699 & 52.79 & \gain{+4.61} & [2.69, 6.57] \\
Pooled control & 819{,}971 & 49.24 & \gain{+8.15} & [6.09, 10.22] \\
\midrule
GAGeo, mask-center swap & -- & 54.22 & \gain{+3.18} & [0.93, 5.44] \\
GAGeo, direct box & -- & 37.16 & +20.24 & -- \\
\bottomrule
\end{tabular}}
\end{table}

\subsection{Qualitative mechanism evidence}
\label{sec:qualitative}
\Cref{fig:qual} separates the two sources of the early lead at 2{,}000 updates. On Tokyo 20282 (\cref{fig:concept}), GAGeo misses the target (IoU 0.28) while the pooled control and our head both hit it (0.73, 0.72): the gain comes from the region-evidence readout, whichever the order. On London 1277 and Paris 1633, matching first lifts the IoU from 0.59 and 0.63 to 0.81. Over all 384 dev queries, matching first is better by more than 0.05 IoU on 186 and worse on 61, and hits 22 queries at IoU 0.5 that merging first misses (9 the other way). The supplementary material follows all 12 prespecified cases along training and adds the four largest final-checkpoint gains (GAGeo IoU ${<}\,0.2$, ours 0.51--0.90). \Cref{fig:mechanism} opens the head on two prespecified cases. The box prompt covers 18--20 query patches, and their coverage weights select the evidence (M2). Before any regression, 91\% of the merged evidence $w$ lies inside the GT box in both cases, and 90\% of it sits on 12 and 27 of the 1{,}369 target patches. Each of these patches votes for a center within half a box of its own (M3), and the expectation of the votes lands 1.8 and 2.5\,px from the GT center in the 518\,px UAV view. These spatial intermediates connect query-region evidence directly to the predicted center.

\section{Conclusion}
Matching before merging turns region evidence into a spatial inductive bias for satellite-to-UAV localization. At equal capacity it adds 8.15\,pp over the pooled control at 2k updates, and the full head needs 5.8--7.0$\times$ fewer updates than GAGeo to reach matched direct-box accuracy. The lead persists at convergence, across all six validation cities, on a held-out city, and against GAGeo's mask-derived boxes. The converged lead is largest at strict IoU thresholds, so the readout also places boxes more precisely.

\newpage
\section*{LLM Usage Statement}
We used AI coding tools, mainly Codex (OpenAI) and Claude Code (Anthropic), to write and run the experiment code, organize the experiment records, generate the figures and tables, check the reported numbers against these records, and discuss how to implement parts of the method. We also used ChatGPT and Claude to search the related work and to draft and polish the text for clarity. The block-style 3D city in \cref{fig:concept}(a) was generated with the image model of ChatGPT; we removed its background and recolored the prompted building. It only illustrates the application scenario: it is not CMA-Loc data, no result is computed from it, and the boxes drawn on it are the measured 2k-update boxes placed relative to the ground truth. All other images in the paper and the supplementary material are CMA-Loc views; no generative model produced them. The ideas, the method, and the conclusions are our own, and we take full responsibility for the content of this paper.

{
    \small
    \bibliographystyle{ieeenat_fullname}
    \bibliography{main}
}


\end{document}